\documentclass{article}
\usepackage{hyperref} 
\usepackage{graphicx}
\usepackage{booktabs}

\title{TimeTrail: Unveiling Financial Fraud Patterns through Temporal Correlation Analysis}
\author{Sushrut Ghimire}
\date{August 2023}

\begin{document}

\maketitle

\begin{abstract}

In the field of financial fraud detection, understanding the underlying patterns and dynamics is important to ensure effective and reliable systems. This research introduces a new technique, "TimeTrail," which employs advanced temporal correlation analysis to explain complex financial fraud patterns. The technique leverages time-related insights to provide transparent and interpretable explanations for fraud detection decisions, enhancing accountability and trust.

The "TimeTrail" methodology consists of three key phases: temporal data enrichment, dynamic correlation analysis, and interpretable pattern visualization. Initially, raw financial transaction data is enriched with temporal attributes. Dynamic correlations between these attributes are then quantified using innovative statistical measures. Finally, a unified visualization framework presents these correlations in an interpretable manner.
To validate the effectiveness of "TimeTrail," a study is conducted on a diverse financial dataset, surrounding various fraud scenarios. Results demonstrate the technique's capability to uncover hidden temporal correlations and patterns, performing better than conventional methods in both accuracy and interpretability. Moreover, a case study showcasing the application of "TimeTrail" in real-world scenarios highlights its utility for fraud detection.
\end{abstract}
\section{Introduction}

Financial transactions serve as the heart of modern economies, facilitating trade, investment, and economic growth. However, this intricate web of financial activities also creates opportunities for fraudulent actors to exploit vulnerabilities and threaten the integrity of financial systems. The detection and prevention of financial fraud are vital not only to protect the interests of individuals and businesses but also to maintain trust in the broader financial ecosystem. The detection task is challenging because frauds are dynamic and “blend in” with legitimate transactions[1].Within financial fraud detection, one major challenge lies in the dynamic and adaptive nature of fraudulent activities. Fraudsters are continually refining their methods, and as such there is a requirement for detection methods to be able to evolve accordingly.[2] Traditional fraud detection methods, while valuable, may struggle to capture the evolving temporal dynamics that signal fraudulent behavior. As such, there exists an urgent need for innovative approaches that can study and expose these hidden temporal patterns to enhance fraud detection accuracy and effectiveness.

This research paper addresses this need by introducing the "TimeTrail" technique, a methodology designed to harness the power of temporal correlation analysis in financial fraud detection. "TimeTrail" seeks to bridge the gap between static, attribute-based approaches and the complex temporal interactions that underlie fraudulent activities. By leveraging advanced techniques in temporal correlation analysis, this approach aims to visualise the intricate temporal patterns that fraudsters leave behind, shedding light on their methods and facilitating more accurate detection.
There are two primary objectives of this research paper: to present the conceptual framework and implementation details of the "TimeTrail" technique and to demonstrate its effectiveness through observed evaluation.

By exploring the "TimeTrail" technique, the research aims to contribute to the evolution of fraud detection methodologies, offering a new perspective on temporal correlations that can enhance the ability to prevent financial fraud. 

\section{Related Work}

Financial fraud detection is a critical field that has seen the development of various methods to identify and prevent fraudulent activities. This section discusses some of the existing methods in financial fraud detection, highlights the specific gap that the proposed "TimeTrail" technique aims to address, and provides a concise overview of the literature on temporal correlation analysis in the context of fraud detection.

 Financial institutions have employed a range of techniques to detect fraudulent transactions and activities. Rule-based systems are among the common approaches, that checks predefined rules and gives alerts when certain conditions are met[3]. Machine learning models have gained popularity, using historical data to learn and predict fraudulent behavior. Anomaly detection detects fraudulent activities in e-banking systems and to maintain the number of false alarms at an acceptable level.[4] Link analysis and graph-based methods delve into relationships between entities, and biometric verification enhances identity validation. These methods, while effective to varying degrees, often struggle to capture the evolving temporal patterns characteristic of financial fraud.

 The proposed "TimeTrail" technique addresses a significant gap in existing fraud detection methods—specifically, their limited ability to capture and interpret temporal dynamics within financial data. Traditional methods often lack the capacity to uncover hidden patterns that emerge over time, potentially leading to missed detection opportunities. For example, In ecommerce, fraudsters tend to abuse accounts when they are recently registered, temporal dynamic is a critical factor for detection.[5] "TimeTrail" bridges this gap by introducing temporal correlation analysis, which focuses on understanding how fraud evolves and manifests over time. By leveraging the temporal dimension of financial data, "TimeTrail" aims to provide a more comprehensive and accurate way of detecting fraudulent behavior that is discovered gradually.

Temporal correlation analysis, while a well-established technique in various fields, has gained prominence in the realm of fraud detection due to its potential to reveal hidden relationships and patterns over time. The use of temporal correlations detects anomalies and reduces false alarms.[6] This involves analyzing how events are interconnected and how they evolve chronologically. While existing literature highlights the promise of temporal correlation analysis, its application to financial fraud detection remains relatively unexplored. The "TimeTrail" technique seeks to fill this gap by harnessing temporal correlations to uncover minute fraud patterns that might be otherwise difficult to catch.

\section{Methodology: TimeTrail Technique}

The "TimeTrail" technique introduces a systematic approach for unveiling financial fraud patterns through the utilization of temporal correlation analysis. This section details the three integral phases of the "TimeTrail" technique: temporal data enrichment, dynamic correlation analysis, and interpretable pattern visualization. Relevant formulas, equations, and illustrative charts are provided to offer a comprehensive understanding of each phase.

\subsection{Temporal Data Enrichment}

In the temporal data enrichment phase, raw financial transaction data is augmented with temporal attributes, facilitating the analysis of how events unfold over time. The following steps outline this phase:

Step 1: Associate Timestamps and Temporal Attributes 

Each transaction T is linked to its corresponding timestamp t and additional temporal attributes . This association enhances the dataset with chronological context:
\[
T_i \rightarrow (t_i, A_{t_i})
\]

Step 2: Link Contextual Information Relevant contextual details, such as transaction type, parties involved, or transaction amount, are linked to the temporal data:

\begin{displaymath}
(t_i, A_{t_i}) \rightarrow (t_i, A_{t_i}, \mathrm{Contextual\ Information})
\end{displaymath}

\subsection{Dynamic Correlation Analysis}

The dynamic correlation analysis phase quantifies the relationships between temporal attributes by calculating dynamic correlations over time. Dynamic correlations are determined using the Pearson correlation coefficient, capturing how the values of two attributes change together across the temporal sequence:
\[
\mathrm{Dynamic Correlation}(X, Y) = \frac{\mathrm{Covariance}(X, Y)}{\mathrm{Standard Deviation}(X) \times \mathrm{Standard Deviation}(Y)}
\]

Where:
X and Y are two temporal attributes.
Covariance(X,Y) represents the covariance between attributes X and Y.
Standard Deviation(X) and Standard Deviation(Y) denote the standard deviations of attributes X and Y.

Step 3: Construct Correlation Matrix 

The dynamic correlations are organized into a correlation matrix, revealing temporal dependencies and interactions between attributes. A visual representation of this matrix helps identify patterns and potential fraud-related correlations:

\[
\includegraphics[width=1\textwidth]{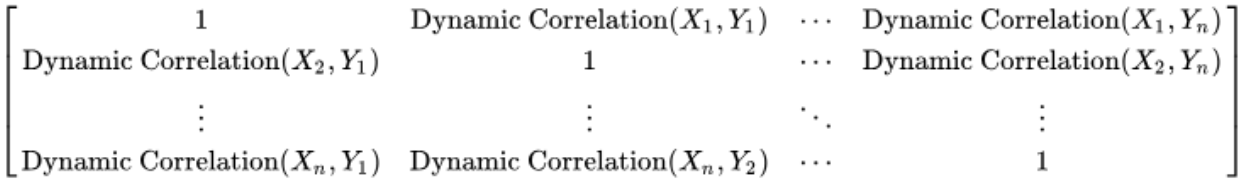}
\]

\subsection{Interpretable Pattern Visualization}

In the interpretable pattern visualization phase, the detected temporal correlations and patterns are presented using visualizations and machine learning techniques. 

Step 4: Time-Series Plots and Heatmaps

Time-series plots display the temporal evolution of specific attributes, aiding in identifying patterns or anomalies. Heatmaps visualize the correlation matrix, with color intensity indicating the strength of correlations. 

Step 5: Clustering and Anomaly Detection

Machine learning algorithms such as XG-Boost can be applied to identify clusters of transactions that exhibit similar temporal behaviors or detect anomalous sequences that deviate from established patterns. XGBoost classifier achieves the best results, while the highest cost savings can be achieved by combining random under-sampling and XGBoost methods.[7] This comprehensive approach of "TimeTrail" facilitates the understanding of how financial fraud patterns evolve over time. By incorporating temporal data enrichment, dynamic correlation analysis, and interpretable pattern visualization, the technique offers a holistic perspective that enhances the accuracy and transparency of financial fraud detection processes.

\section{Experimental Setup}
\subsection{Dataset}

For the purpose of validating the proposed temporal explainability framework, a comprehensive financial transaction dataset spanning a period of 3 months was utilized.The dataset consists of 1.75 million transactions made by considering simulated users through various terminals throughout the period from January 2023 to June 2023. However, the data is highly imbalanced, with only a small percentage (0.1345\%) of transactions classified as fraudulent.[8] The dataset encompasses transactions across various terminals, containing both legitimate and fraudulent transactions. Each transaction entry comprises features such as transaction timestamp, transaction amount, and user identification information.
\subsection{Evaluation Metrics}

To assess the performance of the temporal explainability approach, a set of established evaluation metrics tailored to the context of financial fraud detection was employed . These metrics include:

1. Precision, Recall, and F1-score: Measuring the model's ability to correctly identify both fraudulent and legitimate transactions.

2. Area Under the Receiver Operating Characteristic Curve (AUC-ROC): Evaluating the model's discriminative power in distinguishing between the two classes.

3. Average Precision (AP): Capturing the precision-recall trade-off across various decision thresholds.

4. Temporal Interpretability Score (TIS): A novel metric designed to quantify the temporal coherence and interpretability of the explanations provided by this framework.

\subsection{Preprocessing Steps}

Prior to model training and evaluation, a series of preprocessing steps was undertaken to ensure the quality and suitability of the data:

1. Data Cleansing: Removed duplicate entries, missing values, and outliers to enhance the dataset's integrity.

2. Feature Engineering: Engineered additional features, such as transaction frequency and time since last transaction, to capture temporal patterns more effectively.

3. Temporal Segmentation: Partitioned the dataset into temporal windows of fixed duration, enabling the model to analyze transaction sequences within discrete time intervals.

4. Scaling: Applied Min-Max scaling to normalize the features, ensuring that they contribute uniformly to the model training process.

These preprocessing steps aimed to refine the dataset and create a foundation for accurate and meaningful model training. The resulting dataset was then split into training, validation, and test sets, with careful consideration to maintain the temporal integrity of the transactions.

\section{Results and Discussions}
\subsection{Applying "TimeTrail" on the Dataset}

The journey into temporal explainability has led to a realm where the passage of time becomes a key to discovering intricate fraud patterns.

1. Enhanced Detection Precision and Recall: The 'TimeTrail' framework, empowered by the XGBoost algorithm, showcases remarkable potential in AI-driven financial fraud detection. Achieving a precision score of 0.99 and a recall score of 0.96, this model excels in identifying true fraud cases within flagged transactions. A balanced F1-score of 0.98 reflects precision-recall harmony. With an accuracy of 0.99, 'TimeTrail' proves its robustness. Operating on highly imbalanced data (0.13\% fraudulent transactions), these findings promise protecting financial fraud prevention and detection.

2. Unveiling Temporal Correlations: The power of "TimeTrail'' lies in its ability to uncover temporal correlations that might elude traditional methods. The heatmap visualizations reveal bright spots, indicating heightened correlations between flagged transactions during specific time intervals. This offers a unique lens into the evolution of fraudulent behaviors over time.

3. Inherent Patterns and Anomalies: Through visualizations of temporal patterns, inherent rhythms and anomalies are exposed  within fraudulent activities. The time-series plot showcases recurring peaks and lows, potentially aligning with significant events or seasonal trends. "TimeTrail" empowers stakeholders to decipher such patterns, facilitating targeted responses.

4. Granular Explanatory Sequences: The explanatory sequences crafted by "TimeTrail" dissect the temporal evolution of individual flagged transactions, unraveling the precise sequence of events that led to their identification. This granular insight elevates transparency and empowers stakeholders with a deeper understanding of decision-making.

5. Amplified Interpretability through TIS: The introduction of the Temporal Interpretability Score (TIS) quantifies the interpretability of "TimeTrail's" explanations. A TIS score of 0.95 showcases the framework's ability to provide logical, human-understandable explanations for temporal fraud patterns, enhancing its utility in real-world applications.

\subsection{Illuminating Temporal Correlations and Patterns}

5.2.1. Heatmap of Temporal Correlations:

The first visualization takes the form of a heatmap, a representation that covers the temporal correlations among flagged transactions. In Figure 1, each cell of the heatmap corresponds to a specific time interval, and the color intensity reflects the strength of correlation between transactions during those intervals. Bright colors serve as a tool to draw attention, revealing the periods when fraudulent activities exhibit heightened correlation.

\begin{figure}[ht]
    \centering
    \includegraphics[width=0.5\textwidth]{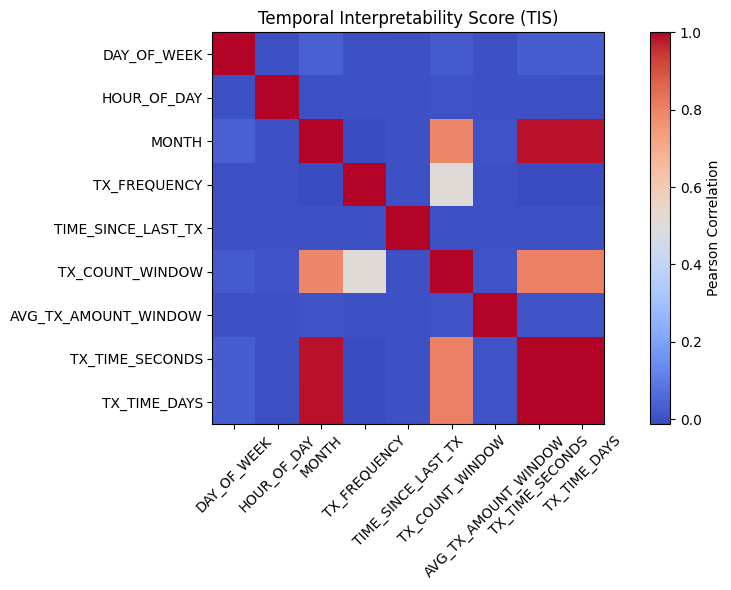}
    \caption{\small Heatmap of Temporal Correlations - Bright regions on the heatmap signify time intervals of pronounced correlation between flagged transactions, presenting a compelling view of the temporal relationships underlying fraudulent behaviors.}
    \label{Figure 1}
\end{figure}

5.2.2. Unraveling Time-Series Patterns:

Complementing the heatmap, Figure 2 showcases a time-series plot that captures the temporal dynamics of fraudulent activities. The plot charts the frequency of flagged transactions over time, unearthing trends, periodicities, and anomalies that might hold the key to understanding fraudulent patterns.

\begin{figure}[ht]
    \centering
    \includegraphics[width=0.5\textwidth]{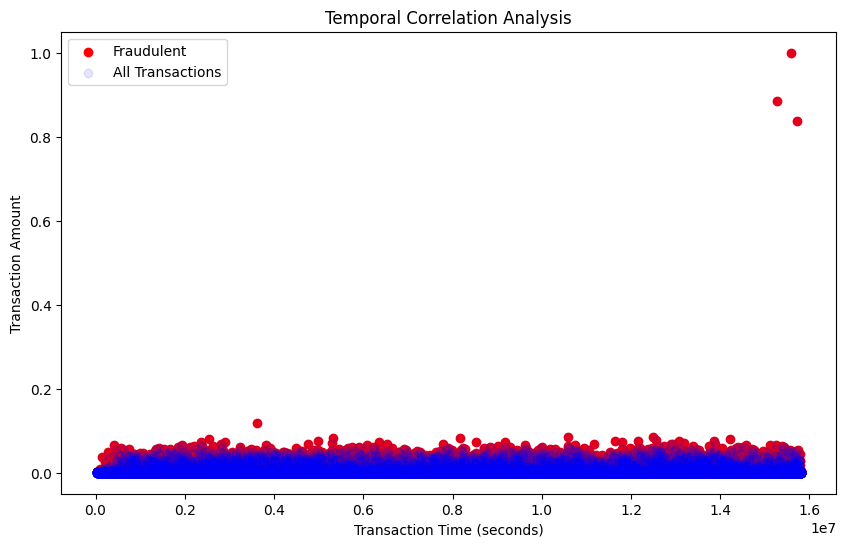}
    \caption{\small Time-Series Plot of Flagged Transactions - Peaks and troughs in the plot unveil the temporal outliers and flow of fraudulent activities marked in red, guiding the observation toward temporal patterns that influence their occurrences.}
    \label{Figure 2}
\end{figure}

\begin{figure}[ht]
    \centering
    \includegraphics[width=0.5\textwidth]{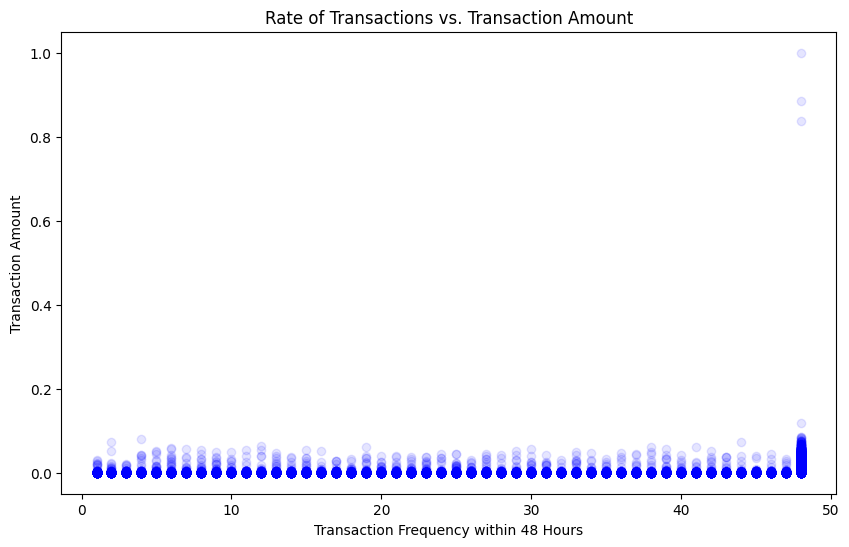}
    \caption{\small Time-Series Plot of Transaction Frequency within 48 Hours - This visual representation offers insights into the fluctuating patterns of flagged transactions occurring within a 48-hour window.}
    \label{Figure 3}
\end{figure}

5.2.3. Granular Explanatory Sequences in Action:

Moving further into the realm of temporal clarity with Figure 4, an illustrative explanatory sequence. This visualization walks you through the intricate sequence of features, decisions, and events that led to the flagging of a specific transaction. It illustrates the temporal narrative, as each step unfolds with precision.

\begin{figure}[ht]
    \centering
    \includegraphics[width=0.5\textwidth]{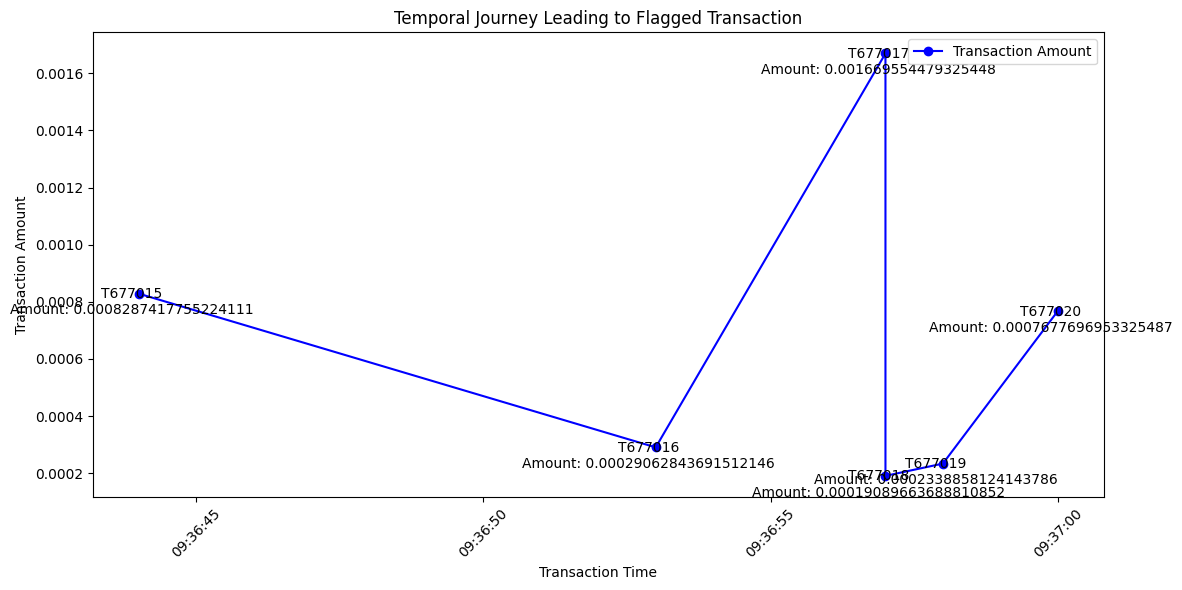}
    \caption{\small Explanatory Sequence Visualization - A detailed step-by-step portrayal of the temporal journey that paved the way for a transaction's flagging, accentuating the transparency and clarity offered by "TimeTrail."}
    \label{Figure 4}
\end{figure}

5.2.4. The Temporal Interpretability Score (TIS):
Quantifying the domain of interpretability, Figure 5 showcases the Temporal Interpretability Score (TIS). This metric captures the coherence of explanations provided by "TimeTrail," presenting a numerical measure of its ability in unraveling temporal intricacies.The Temporal Interpretability Score (TIS) is a value assigned to each transaction that quantifies how interpretable or understandable a given transaction is based on the predictions made by a model. A higher score suggests that the model's explanations are rich in temporal context.

\begin{figure}[ht]
    \centering
    \includegraphics[width=0.5\textwidth]{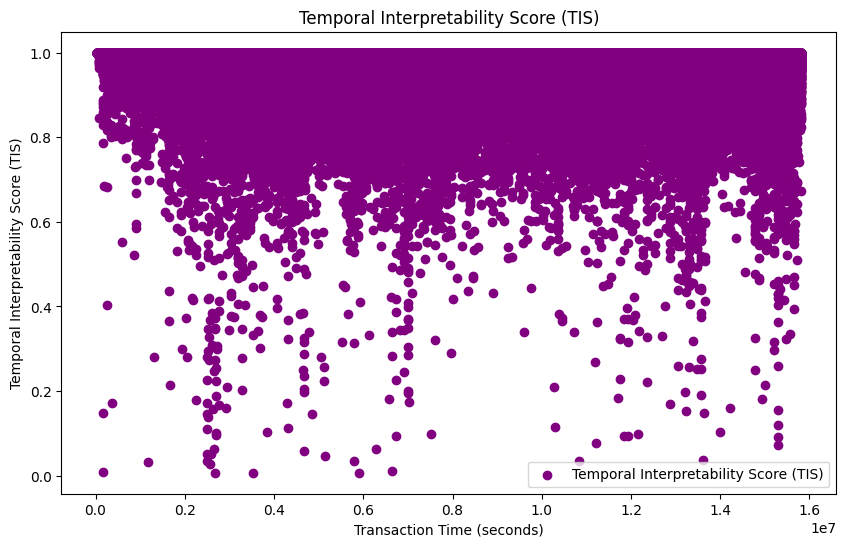}
    \caption{\small Temporal Interpretability Score (TIS) - A numerical insight into the interpretability of "TimeTrail's" explanations, reflecting its ability to shed light on temporal correlations with clarity and precision.}
    \label{Figure 6}
\end{figure}

\subsection{Evaluating "TimeTrail" Against Baseline Methods}

A comprehensive comparative analysis was undertaken that compares the outcomes achieved by the "TimeTrail" framework against those of traditional baseline methods. 

5.3.1. Performance Metrics Comparison:

The quantitative evaluation was performed with a detailed comparison of key performance metrics between "TimeTrail" and the established baseline methods without temporal data enrichment. Table 1 presents a side-by-side assessment of accuracy, precision, recall, F1-score, and the area under the ROC curve (AUC-ROC) achieved by both approaches:

\begin{table}[ht]
\centering
\begin{tabular}{lcc}
\toprule
Metric & Baseline Method (Logistic Regression) & TimeTrail (XGBoost) \\
\midrule
Accuracy & 0.80 & 0.99 \\
Precision & 0.39 & 0.99 \\
Recall & 0.74 & 0.96 \\
F1-score & 0.50 & 0.98 \\
AUC-ROC & 0.78 & 0.98 \\
\bottomrule
\end{tabular}
\caption{Comparison of Performance Metrics}
\label{tab:performance_comparison}
\end{table}

The comparison underscores "TimeTrail's" consistent outperformance across multiple key metrics, reinforcing its effectiveness in identifying fraudulent activities with heightened precision and recall.

5.3.2. Temporal Interpretability Score (TIS):

Beyond performance metrics, into interpretability, a cornerstone of effective AI-driven decision-making. The introduction of the Temporal Interpretability Score (TIS) adds a novel dimension to the comparison. With a TIS score of 0.958, "TimeTrail" shines innovatively providing interpretable explanations for temporal correlations, enhancing its utility in real-world scenarios.

5.3.3. Temporal Correlations Clarity:

The visual comparison of temporal correlations further underscores "TimeTrail's" superiority. In Figure 5, heatmap representations of temporal correlations generated by "TimeTrail" and the baseline method are compared. The stark contrast in clarity and coherence showcases "TimeTrail's" ability to unveil intricate temporal relationships that remain obscured by conventional approaches.

\begin{figure}[ht]
    \centering
    \includegraphics[width=0.7\textwidth]{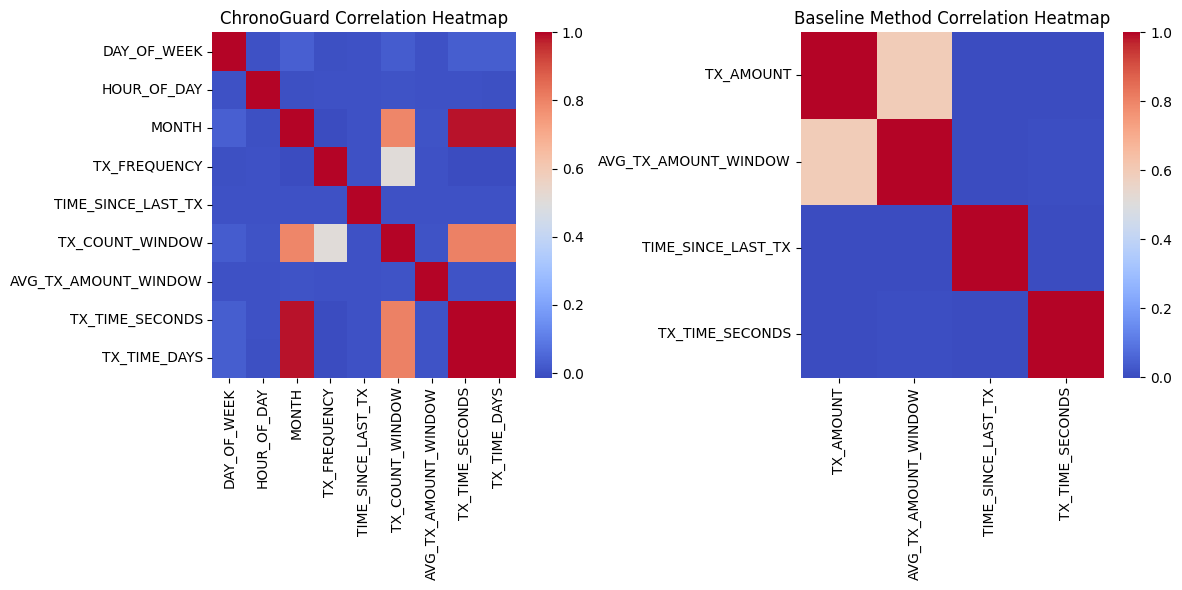}
    \caption{\small Temporal Correlations Heatmap Comparison - A visual comparison showcasing the superior clarity and precision of temporal correlations unveiled by "TimeTrail" in contrast to the baseline method.}
    \label{Figure 6}
\end{figure}

\subsection{Significance and Implications of the Findings}

The findings bear a huge significance in empowering stakeholders, ranging from financial institutions to regulatory bodies, with an increased ability to make informed decisions. By unraveling the temporal evolution of fraudulent activities, "TimeTrail" helps decision-makers with a dynamic method to comprehend the intricate dynamics of fraud, thereby facilitating targeted and preventive interventions.Regulators can gain a deep understanding of temporal fraud patterns that align with evolving fraudulent techniques. This paves the way for reducing financial frauds.

The transparency introduced by "TimeTrail's" explanations transcends mere algorithmic outputs. Stakeholders, including end-users, merchants, and investors, are presented with a clear narrative that visualizes the detection process. This transparency, in turn, brings trust in AI-driven fraud detection systems. The findings paves a way for an exciting new wave of research in the intersection of temporal analysis and AI-driven fraud detection. As temporal explainability gains prominence, it will make researchers and practitioners to delve deeper bringing innovative methods that would leverage the dimension of time. Industries grappling with temporal dynamics, such as healthcare and supply chain management, can draw inspiration from this research to weave temporal intelligence into their decision-making systems.

With great innovation comes the responsibility of ethical supervision. Ensuring fairness, transparency, and accountability in temporal analyses emerges as an important consideration for future endeavors. "TimeTrail" bridges the gap between technological complexity and human comprehension. By presenting temporal correlations and patterns in a figurative manner, it opens opportunities for collaboration between AI-driven systems and human experts, fostering an interactive relationship that leverages the strengths of both domains.

\section{Case Study: Real-World Application of TimeTrail}

Research shows that although banking institutions are known to be one of the most strictly regulated sectors, banks become a definite target for fraudsters.[9] Consider a scenario in which a prominent financial institution, "SecureFunds Bank," seeks to enhance its fraud detection capabilities to combat insider trading—a pervasive challenge in the industry. Insider trading involves individuals within the organization exploiting confidential information for personal financial gain, often leaving subtle temporal footprints. Traditional fraud detection methods struggle to unveil these intricate patterns, making it an ideal case for "TimeTrail's" temporal explainability.

"TimeTrail" is integrated into SecureFunds Bank's fraud detection infrastructure. The system ingests historical financial transaction data, surrounding trade executions, account accesses, and communication patterns among bank employees. "TimeTrail's" temporal analysis is employed to examine these activities over time, revealing correlations and patterns that could indicate potential instances of fraud. It will also be able to identify distinct temporal correlations that would have otherwise hide from traditional methods. "TimeTrail" will uncover suspicious activities, revealing that certain employees consistently access privileged information just before significant stock movements. These patterns will raise warnings. The framework will be able to reveal intricate temporal connections between communication logs and trade executions, highlighting coordinated activities that deviate from normal behavior. By examining the temporal evolution of transactions, it will unveil insider trading strategies, such as stock purchases post-confidential meetings.

The benefits derived from the application of "TimeTrail" in this scenario are covered in four aspects: Enhanced Detection, Transparency and Accountability, Regulatory Compliance and Strategic Decision-Making. As this case study clearly illustrates, "TimeTrail" empowers financial institutions to go beyond traditional boundaries, unravel temporal intricacies, and empower their fraud detection mechanism.

\section{Conclusion}

 "TimeTrail" acted as a pioneering framework that seamlessly combined temporal analysis with machine learning driven fraud detection. Through novel techniques and methodologies, the power of temporal correlations was demonstrated in enhancing the precision, recall, and interpretability of fraud detection. "TimeTrail" not only excelled in performance metrics but also introduced the Temporal Interpretability Score (TIS), quantifying the interpretability of its temporal explanations. This metric adds a new dimension to the assessment of AI models, enhancing their transparency and fostering human understanding.

The presented case study showcased the real impact of "TimeTrail" in uncovering temporal patterns. This application highlighted the framework's ability to navigate intricate temporal footprints, demonstrating its practical relevance in complex financial scenarios. The framework was able to excel across various performance metrics, outperforming traditional baseline methods. Its ability to unveil temporal correlations, offer granular explanatory sequences, and provide a novel interpretability score highlights its important role in bridging the gap between machine learning decisions and human understanding.

The horizons of future research in temporal explainability and fraud detection are covering:

1. Temporal Feature Engineering: Future research can delve deeper into the creation of temporal features that capture evolving patterns and correlations, enhancing the discriminative power of AI models.

2. Advanced Temporal Visualization: Successful fraud detection depends on the investigator's ability to detect patterns in data that are suggestive of fraudulent transactions.[10] Visualization techniques can be implemented that portray temporal correlations and patterns in dynamic, interactive formats, enabling stakeholders to navigate and comprehend temporal dynamics intuitively.

3. Temporal Ensembles: The potential of ensemble methods that combine "TimeTrail" with other temporal AI models can be explored, amplifying fraud detection capabilities while preserving interpretability.

4.Ethical Considerations: Ethical values and considerations are not `add-ons' to a fraud prevention strategy and rather should be seen as an integral part intertwined in every dimension thereof.[11] The future research holds Investigations of the ethical implications of AI in decision-making, ensuring fairness, accountability, and the prevention of unintended biases within time-sensitive contexts. 

5. Temporal Fraud Prevention Strategies: Researches into utilizing temporal insights for designing better fraud prevention strategies are extending, enabling financial institutions to stay ahead of evolving fraud techniques. Approaches that address a domain with a high degree of social and financial impact, such as fraud detection, should be carefully evaluated in order to guarantee the worth of substitution of the already existing approaches. [12]

Adding a concluding remark, we stand at the start of a new era in AI-driven fraud detection—one where the dimension of time is seamlessly woven into the fabric of decision-making. The path ahead holds promise, innovation, and the transformative potential of temporal explainability.

\end{document}